\documentclass[conference]{IEEEtran}
\IEEEoverridecommandlockouts

\usepackage{cite}
\usepackage{amsmath,amssymb,amsfonts}
\usepackage{algorithmic}
\usepackage{graphicx}
\usepackage{textcomp}
\usepackage{xcolor}
\usepackage{url}
\usepackage{authblk}
\def\BibTeX{{\rm B\kern-.05em{\sc i\kern-.025em b}\kern-.08em
    T\kern-.1667em\lower.7ex\hbox{E}\kern-.125emX}}

\begin{document}

\title{Data-Driven Dynamic Algorithm Dispatch with Large Language Models\\

\thanks{This material is based upon work supported by the Defense Advanced Research Projects Agency (DARPA) under Agreement No. HR00112490488.
}
}

\author[1,2]{Rushil Shah}
\author[1,3]{Emmanuel Lujan}
\author[1,4]{Rabab Alomairy}
\author[1,5]{Alan Edelman}

\affil[1]{Computer Science \& Artificial Intelligence Laboratory,  Massachusetts Institute of Technology, USA.}
\affil[2]{
\textit {rshah2@mit.edu}
\quad$^{3}$\textit {eljn@mit.edu}\protect 
\quad$^{4}$\textit {rabab.alomairy@mit.edu}
\quad$^{5}$\textit {edelman@mit.edu}\protect\\
}

\maketitle

\begin{abstract} 
We introduce a large language model (LLM)-driven approach for generating dynamic algorithmic dispatch heuristics in high-performance linear algebra. By combining prompt engineering with LLaMA 3 and a curated performance database, the model learns to synthesize selection heuristics that exploit structural patterns to identify fast algorithmic choices. A case study on LU factorization demonstrates the model’s ability to replicate expert-designed strategies. This work, developed as part of the DARPA–MIT SmartSolve project, highlights the promise of LLMs for algorithmic discovery and the development of more adaptive, fast linear algebra software.
\end{abstract}


\section{Introduction}

Specialized variants of classical linear algebra algorithms have been developed to exploit the structural patterns of the input data, often yielding significant performance improvements. A key example is matrix factorization, where selecting an appropriate strategy can critically impact the efficiency of solving linear systems, computing eigenvalues, and performing statistical estimation. Among these, LU decomposition is widely used for its ability to reduce problems to triangular solves. Variants such as banded LU and KLU~\cite{klu2010} are explicitly designed to leverage data properties such as sparsity and block patterns, achieving substantial speed-ups and memory savings over dense or general-purpose approaches.

Current numerical software—including those in Julia, Python, and MATLAB—incorporate built-in heuristics to guide algorithm dispatch. These rules offer many opportunities for optimization as they are increasingly challenged by the growing diversity of matrix structural patterns and solver designs~\cite{Li2005-bz}. As research continues to introduce specialized algorithms targeting different matrices, it becomes increasingly difficult to rely on manually engineered heuristics.

To address this challenge, the DARPA-MIT SmartSolve project \cite{SmartSolve2025} proposes a Julia-based framework aimed at accelerating computations by producing dynamic dispatch heuristics that optimize both algorithmic and architectural choices.  The SmartSolve methodology, as applied to linear algebra, starts with a discovery process that systematically evaluates various algorithms—such as the aforementioned banded LU and KLU—across a diverse set of matrix patterns from Matrix Depot~\cite{zhang2016matrix}, such as Hilbert, Vandermonde, and Toeplitz, along with multiple data formats, mixed-precision strategies, and computer architectures. For each combination, a variety of metrics are collected, including features of the matrix pattern—such as dimension, sparsity, and condition number—as well as performance metrics like type conversion overhead, computational runtime, and numerical accuracy. These measurements are aggregated into a performance database. Then, an automated Pareto analysis identifies optimal trade-offs between speed and precision [6]. The resulting database is used to train a data-driven model that generates a heuristic for selecting the optimal combination of algorithmic and architectural choices based on the input matrix. These newly developed heuristics aim to improve the performance of modern linear algebra libraries, such as \texttt{LinearSolve.jl}, which provides fast implementations of linear solving algorithms in Julia.

Large language models (LLMs) can serve as a powerful mechanism for recognizing patterns in benchmarking data and synthesizing decision logic. Prior initiatives like ChatHPC have shown that LLMs can generate performant numerical code and even propose novel optimization strategies [4, 5].  
More recently, Google Deepmind released AlphaEvolve \cite{novikov2025alphaevolvecodingagentscientific}, an LLM agent used to discover and optimize general purpose algorithms. It devised a method for multiplying 4×4 complex matrices with just 48 scalar multiplications—one fewer than Strassen’s 1969 algorithm, marking the first known improvement in over half a century.

Here, we present our latest advances in the SmartSolve project: an LLM-driven approach for dynamic algorithmic dispatch. 
 
Our main contributions are as follows: (1) An LLM-driven approach for automatic generation of dynamic algorithm dispatch heuristics tailored to computational linear algebra.
 
(2) A case study using LLaMA 3 to rediscover a dispatch heuristic for LU factorization applied to a wide range of matrices provided by \texttt{MatrixMarket.jl}. Through these contributions, we aim to extend the \texttt{SmartSolve.jl} framework beyond traditional machine learning-based strategies.

\section{Methods}
\label{Methods}

\begin{figure}[htbp]
\centerline{\includegraphics[width=0.5\textwidth]{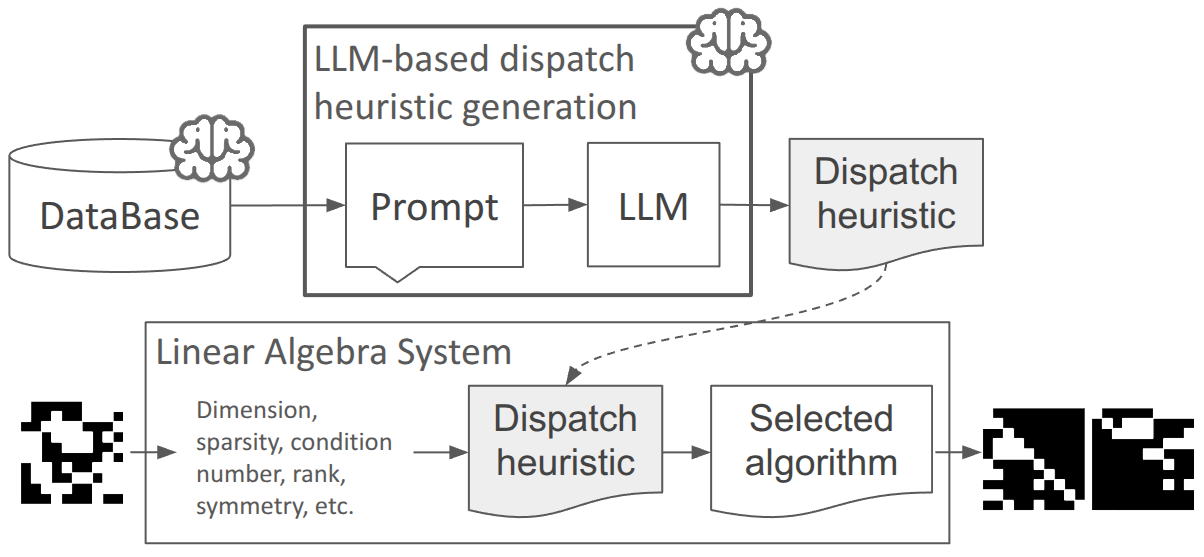}}
\caption{Overview of our LLM-driven approach for generating dynamic algorithm dispatch heuristics in linear algebra.}

\label{fig:main-diagram}
\end{figure}

Our approach, as illustrated in the top portion of Fig.~\ref{fig:main-diagram}, leverages prompt engineering to instruct the LLM to generate dynamic algorithm dispatch heuristics. 

The prompt is designed following three general guidelines:
(1) The model is first provided with a clear description of its role and objective. For instance, instructing the LLM to act as an expert in matrix factorization helps constrain its reasoning to the relevant domain and fosters more specialized responses.
(2) Include \texttt{SmartSolve.jl}’s performance database—which contains benchmark results for a variety of algorithms—as part of the prompt. This endows the model with information beyond its original training corpus and enables it to learn correlations between input matrix features and algorithmic performance, thereby facilitating the generation of informed dispatch heuristics.
(3) Request the LLM to generate the desired heuristic using the database according to a defined output format. The LLM is instructed to produce a tree-based dispatch algorithm expressed using \texttt{if}/\texttt{else} statements to enhance human interpretability. This is essential to mitigate variability in LLM outputs, which may otherwise include inconsistent formatting or reference algorithms outside the provided dataset. 

Finally, as illustrated in the bottom part of Fig.~\ref{fig:main-diagram}, the resulting heuristic maps the features of the input matrix to the most efficient algorithm available. This heuristic is ultimately intended for integration into modern linear algebra systems, such as Julia’s \texttt{LinearSolve.jl}. 

\section{Results and Discussion}

\begin{figure}[htbp]
\centerline{\includegraphics[width=0.40\textwidth]{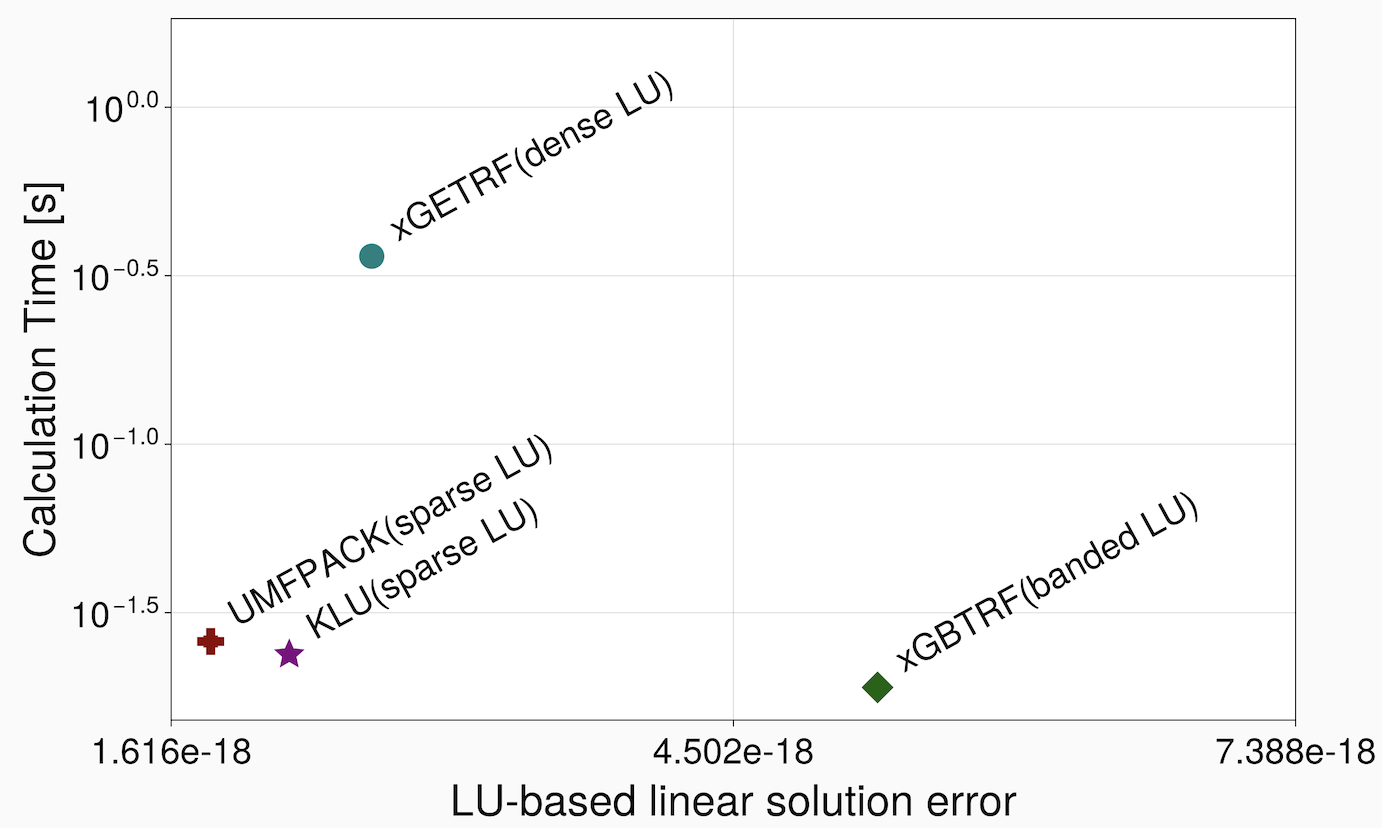}}
\caption{Time and accuracy trade-offs of LU performance on a $2^{12}$x$2^{12}$ Poisson matrix, used to guide LLM-generated dispatch heuristics.}
\label{fig:pareto}
\end{figure}

Our case study focuses on rediscovering one of \texttt{SmartSolve}’s LU-based dynamic dispatch heuristics using our LLM-driven approach.  
\texttt{SmartSolve.jl} is first used to generate a comprehensive performance database that benchmarks multiple LU factorization strategies—including dense, sparse, and banded solvers—on structurally diverse matrices. An automated Pareto analysis is then applied to identify configurations that achieve optimal trade-offs between runtime and numerical accuracy.  As shown in Fig.~\ref{fig:pareto}, the dense LU algorithm with partial pivoting (\texttt{xGETRF}) exhibits higher computational cost. By converting the matrix to a sparse representation and leveraging specialized solvers—such as the Unsymmetric MultiFrontal Method (\texttt{UMFPACK} ), \texttt{KLU} , or the banded LU factorization (\texttt{xGBTRF})—comparable accuracy is attained at a fraction of the computational cost.
 
The resulting Pareto-optimal choices are then used to guide the LLM in generating the desired heuristics. The results presented here were obtained using double precision computations.

The prompt was constructed following the guidelines outlined in the \ref{Methods} section. The portion corresponding to guideline (1) and (3)—which provides contextual information for the task and the output constraints—is potentially broadly applicable to a wider class of algorithms, including QR decomposition, SVD, and FFT. In contrast, the portion related to guideline (2) is specific to the performance database generated for the target algorithm.
This case study is implemented in a Julia-based notebook that leverages 7B Mistral v0.3 through Ollama v0.9.4. The code is publicly available in the \texttt{SmartSolve.jl} GitHub repository \cite{SmartSolve2025}.

Our results demonstrate the successful rediscovery of the desired LU dispatch heuristic. However, while the model is capable of producing meaningful decision logic, its inherently statistical nature can lead to variability across outputs. As such, iterative refinement—such as prompt rephrasing or multiple generation rounds—is often required to obtain reliable results. Another point to consider is that prompt engineering is constrained by limited input space of the model’s pretrained knowledge. Fine-tuning could mitigate these limitations by adjusting the model’s internal parameters to internalize a large-scale performance database, offering a more scalable solution. 

This study provides evidence for the viability of LLMs in generating dynamic algorithmic dispatch heuristics, while highlighting key limitations—namely, variability from their probabilistic nature and prompt size constraints.

We expect to influence the design of faster linear algebra software and catalyze innovation in frameworks such as \texttt{SmartSolve.jl}.

\bibliographystyle{abbrv}
\bibliography{biblio}

\end{document}